\documentclass[letterpaper, 10 pt, conference]{ieeeconf}  

\IEEEoverridecommandlockouts
\usepackage{graphicx}
\usepackage{amsmath,amssymb,mathtools} 
\usepackage{multirow}
\usepackage{tabularx}
\usepackage{booktabs}
\usepackage{hyperref}
\usepackage[misc]{ifsym}

\usepackage[colorinlistoftodos]{todonotes}
\usepackage{verbatim}

\title{\LARGE \bf
Robust Moving Objects Detection in Lidar Data\\ Exploiting Visual Cues
}

\author{Gheorghii Postica$^{1}$ Andrea Romanoni$^{1}$  Matteo Matteucci$^{1}$
\thanks{$^{1}${Politecnico di Milano, Dipartimento di Elettronica, Informazione e Bioingegneria (DEIB),  Milano, Italy}
		\vfill{\tt\footnotesize george.postica@mail.polimi.it} 
        \vfill{\tt\footnotesize andrea.romanoni@polimi.it (corresponding author)}
        \vfill{\tt\footnotesize matteo.matteucci@polimi.it}
        }
        }%

\begin{document}

\maketitle
\thispagestyle{empty}
\pagestyle{empty}

\begin{abstract}
Detecting moving objects in dynamic scenes from sequences of lidar scans is an important task in object tracking, mapping, localization, and navigation. 
Many works focus on changes detection in previously observed scenes, while a very limited amount of literature addresses moving objects detection.
The state-of-the-art method exploits Dempster-Shafer Theory to evaluate the occupancy of a lidar scan and to discriminate points belonging to the static scene from moving ones.
In this paper we improve both speed and accuracy of this method by discretizing the occupancy representation, and by removing false positives through visual cues.
Many false positives lying on the ground plane are also removed thanks to a novel ground plane removal algorithm. Efficiency is improved through an octree indexing strategy.
Experimental evaluation against the KITTI public dataset shows the effectiveness of our approach, both qualitatively and quantitatively with respect to the state-of-the-art.
\end{abstract}

\section{Introduction}
\label{sec:intro}
Moving object detection has been acknowledged to be a crucial step in many applications (e.g., autonomous driving, advanced driver assistance systems, robot navigation, video surveillance, etc.) where specific targets such as  people, vehicles, or animals, have to be detected before operating more complex processes.
In robotics the observer is moving while it operates in the environment, and it becomes hard to distinguish which object is moving with respect to the static scene due to egomotion effects; this affects all sensors used in mobile robotics being these laser range finders or cameras.

An example of moving object detection in laser data is the work by Azim and Aycard~\cite{azim2012detection}; in their work, the authors propose to store perceived point clouds in an octree-based occupancy grid, and look for inconsistencies between subsequent scans. Each voxel (octree cell) of the occupancy grid is classified as free or occupied through ray tracing; voxels classified as both occupied and free in different scans, are called as dynamic. 
Dynamic voxels are then clustered and filtered such that clusters whose bounding box shape differs significantly from fixed size boxes, are removed. In the authors scenario, fixed sized boxes represent cars, trucks and pedestrians, therefore, the approach was targeted at a specific set of objects.

The former example is one of the few cases of laser-based moving objects detection algorithm. Indeed an extended laser-based literature focuses on the closely related, and possibly simpler, change detection problem~\cite{vieira2014spatial,andreasson2007has,drews2013fast,xiao2013change}. 
Change detection  aims at detecting changes in an observed scene with respect to a previously stored map of the environment, e.g., to understand if an object appears or disappears. 
Conversely, in moving objects detection, the map is unknown a-priori and the moving objects can only partially disappear, i.e., between two consecutive observations a region of a moving object remains occupied, therefore appearing as a static item.

Andreasson \emph{et al.}~\cite{andreasson2007has} and Nu{\~n}es \emph{et al.}~\cite{nunez2010change} represent laser scans through a set of distributions, respectively the Normal Distribution Transform and the Gaussian Mixture Model, to detect changes where the distributions differ significantly. 
Vieira \emph{et al.}~\cite{vieira2014spatial} cluster the laser points into implicit volumes an through Boolean operators detect the regions of change.
Xiao \emph{et al.}~\cite{xiao2013change} model the physical scanning mechanism using Dempster-Shafer Theory (DST), and provide sound statistical tools to evaluate the occupancy of a scan and to compare the consistency among scans, i.e., to detect the moving object. 
Our contribution, in this paper, is inspired by the latter work.

As far as cameras are concerned, classical image-based methods to detect moving objects in a video sequence are based on the difference between a model of the background, i.e., the static scene, and the current frame (see \cite{Piccardi2004background} and \cite{Sobral2014}). 
Such algorithms require the camera to be fixed. 
Some extensions are able to handle jittering or moving cameras by registering the images against the background model \cite{azzari2005effective,Andrea,kim2013detection,shakeri2014detection}.
However this class of algorithms needs information about the appearance of the background scene and in most cases, e.g., with a surveying vehicle, this assumption does not hold. 
Other approaches cluster optical flow vectors \cite{markovic2014moving}, or rely on deep learning \cite{lin2014deep}.

Laser range finders and cameras have complementary features; the former are able to provide 3D 360-degree accurate measurements of the environment, the latter capture the appearance of the environment. Only few authors proposed hybrid approaches to combine laser data with the visual information provided by a camera in moving object detection.
Premebida \emph{et al.} \cite{premebida2009lidar} proposed to join two classifiers based on laser camera features to detect pedestrians moving in front of the observer; in this case the scope was limited and the proposed algorithm would need a not trivial extension of the training to deal with general moving objects.
Vallet \emph{et al.} \cite{vallet2015extracting} extended the change detection algorithm presented by Xiao \emph{et al.} in \cite{xiao2013change} to detect moving objects. 
Moreover they exploit visual information by projecting into the image the laser 3D points and by segmenting the moving objects through a graph cut algorithm that takes into account laser label consistency, a smoothness term, and a penalization in the labeling where the image shows edges. 

In this paper, we propose a novel hybrid approach to improve the accuracy of state-of-the-art laser-based moving objects estimation and speed up its computation thanks to a novel ground plane detection algorithm and octree representation; in addition we propose an image based validation test to diminish false positives detection. 
Section \ref{sec:lidar} introduces the novel laser-based moving objects detection method. In Section \ref{sec:images} we show how we improve its robustness against false positive detection by exploiting image information.
In Section~\ref{sec:experiments} we illustrate the results of our algorithm over the KITTI~\cite{Kitti} public dataset, and in Section \ref{sec:concl} we provide some insights on future developments in the paper conclusion.

\section{Laser-based Moving Objects Detection}
\label{sec:lidar}
\begin{figure}[t]
\centering
\includegraphics[width=0.9\columnwidth]{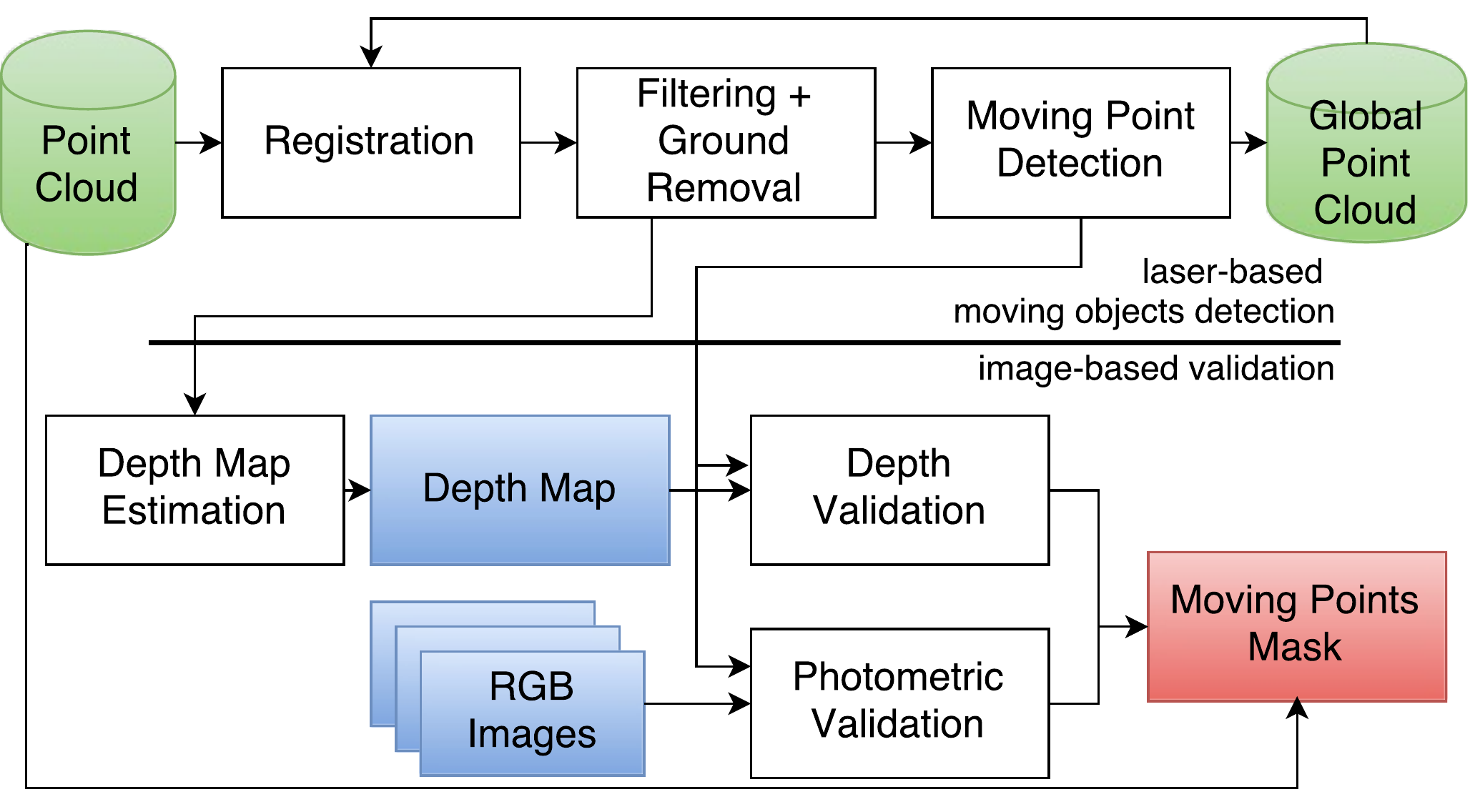}
\caption{Moving Object Detection process.}
\label{fig:algo}
\end{figure}

In the following we focus on the laser-based moving object detection setting in which we process a sequence of 3D point-clouds incrementally; as an example, consider a Velodyne lidar on the top of a car moving in a urban area with the aim of building a map of it. 
We keep a model of the static scene, initialized from the first point cloud, in the form of a 3D map and we update it by fusing subsequent point clouds after dynamic objects removal. 
The reference pipeline for this task is depicted in the upper part of Fig.~\ref{fig:algo} where, in the filtering block, we include also the novel ground plane removal algorithm.

\subsection{Point cloud registration and filtering}
As a new point cloud is generated by the laser range finder, we align it to an existing map, initialized with the first scan, through the Generalized Iterative Closest Point (GICP) algorithm~\cite{segal2009generalized}.
After point cloud alignment, we remove the points having a distance from the point cloud center greater than a given threshold $\tau = 30m$, along any of the three main axis to neglect points too faraway from the sensor.

Directly adding the aligned points to the map would lead to a very dense result, with possibly repeated points; instead, we compare the new point cloud with the last $W = 10$ 
point clouds we recently aggregated, and we add each point only if no other close point exists already in the map. This fills the gaps in point clouds and makes the global cloud free of duplicates\footnote{The use of the term duplicate, in this context, is improper since it is very unlikely the lidar samples exactly the very same point, but, assuming the sampling beam has non negligible size, we have overlapping regions sampled repeatedly and this would induce an unnecessary oversampling of the environment.}. Once the new points have been selected for addition, we further simplify the point cloud by ground plane removal.

\subsection{Ground plane removal}
\begin{figure}
\includegraphics[width=\columnwidth]{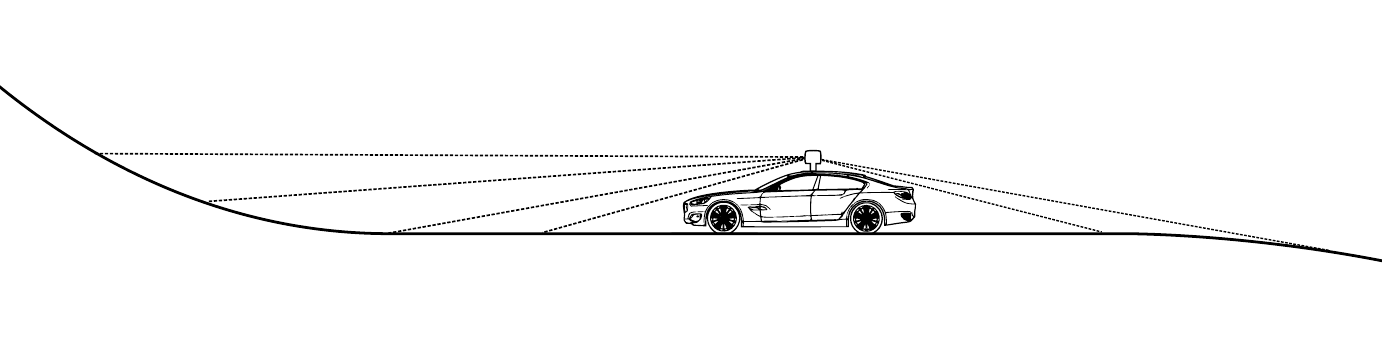}
\caption{Non trivial example in which naive and plane fitting based ground removal fail.}
\label{fig:nonplane}
\end{figure}

Subsequent laser measurements that lay on the ground plane convey redundant and negligible information about moving objects since the ground plane is expected to be mostly static. Therefore, as a further filtering step, we classify and remove ground plane points. 
A naive approach to do that would discard all points which are under a certain negative height from the laser sensor. 
A slightly better approach fits a horizontal plane, e.g., with RANSAC, and removes points which lay on it.
The drawback of both approaches arises whenever we deal  with a non-planar ground surface, as in Fig.~\ref{fig:nonplane}, or errors in extrinsic sensor calibration.

\begin{figure}[t]
\centering
\includegraphics[width=0.4\columnwidth]{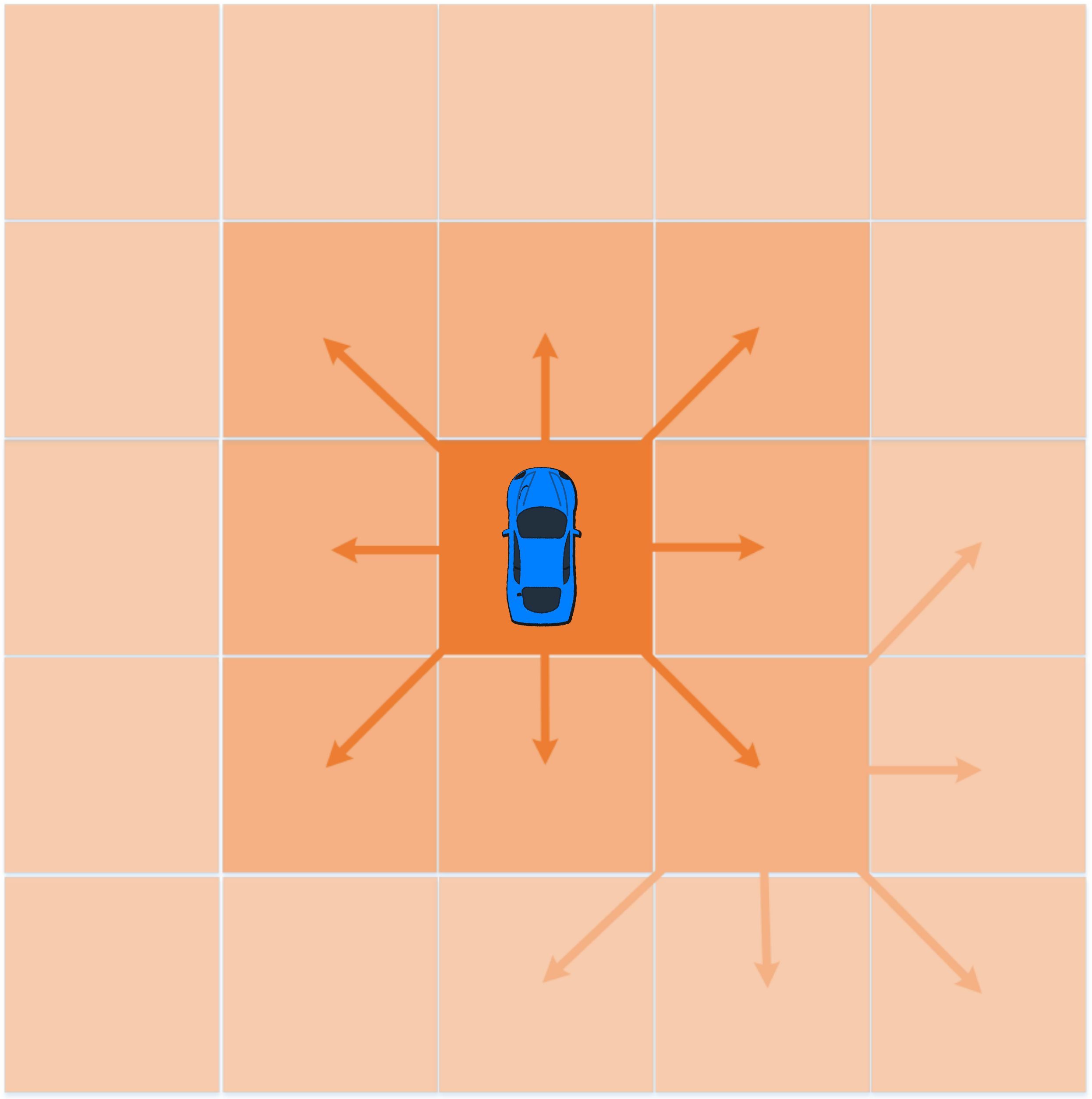}
\caption{Schema of ground height propagation.}
\label{fig:propag}
\end{figure}

\begin{figure}
\setlength{\tabcolsep}{1pt}
\begin{center}
\begin{tabular}{cc}
\includegraphics[width=0.48\columnwidth]{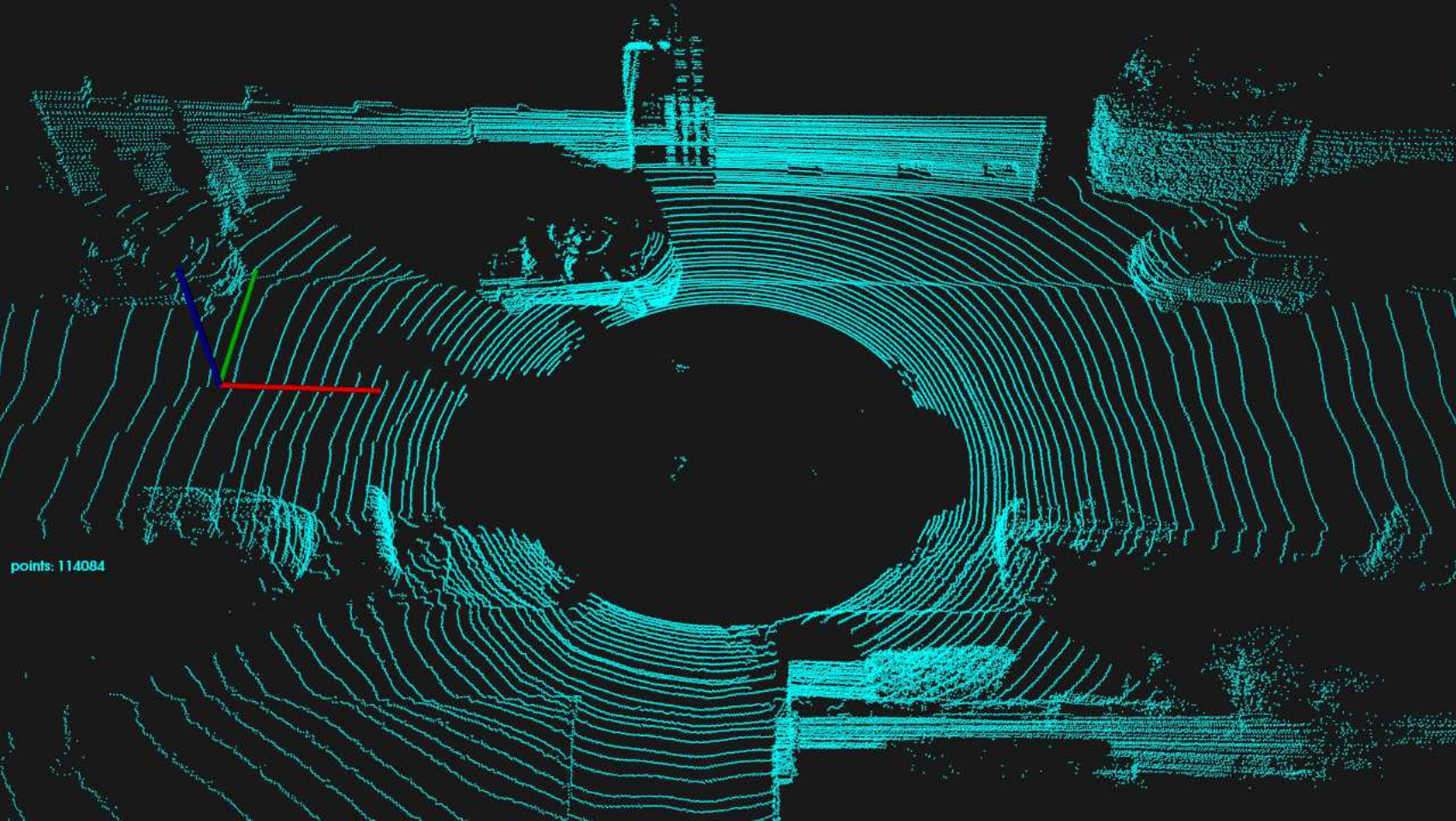} &
\includegraphics[width=0.48\columnwidth]{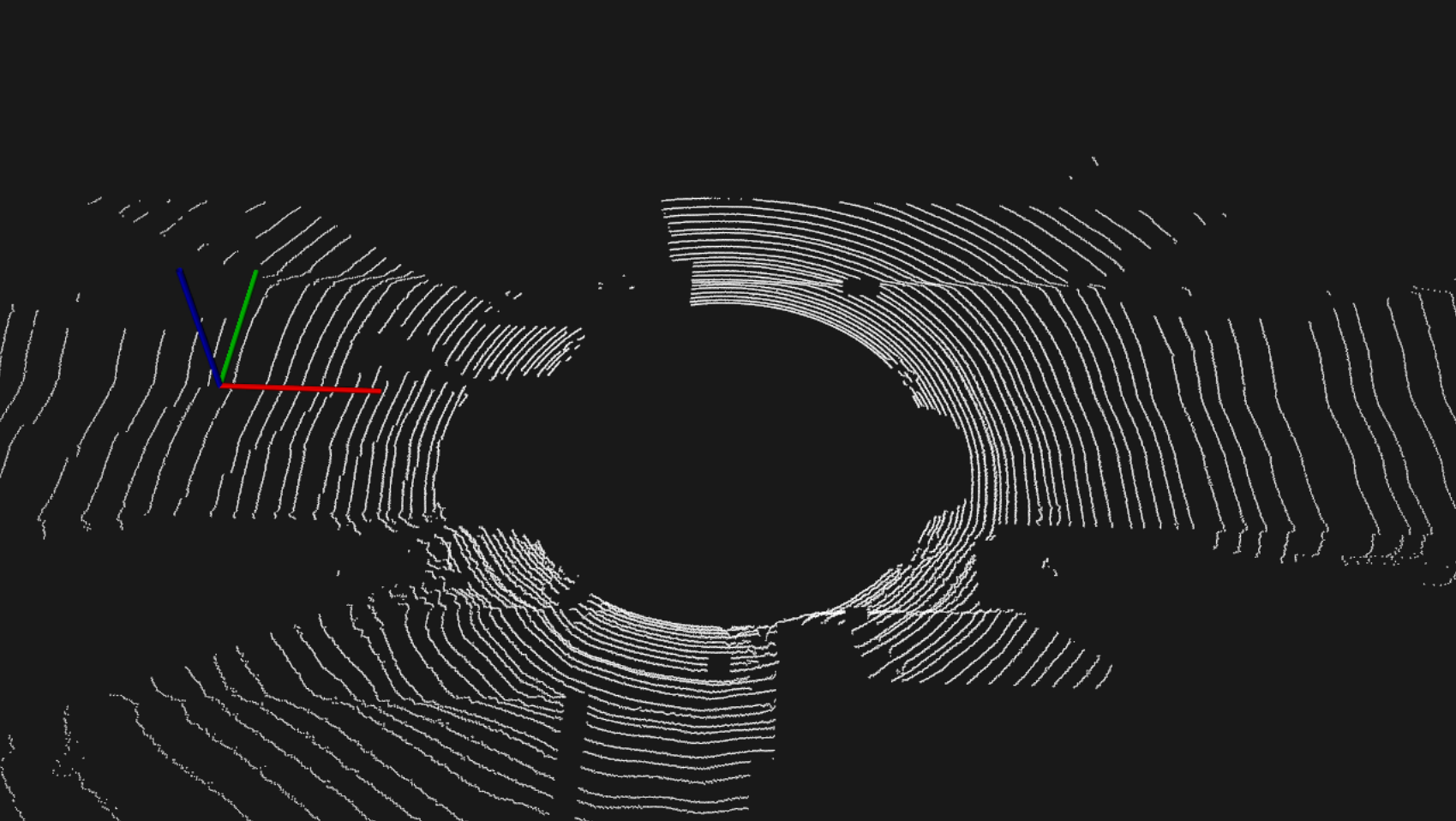}\\
\end{tabular}
\end{center}
\caption{A scan (left) and the ground points to be removed (right).}
\label{fig:groundremoval}
\end{figure}

In this paper we propose to remove the unnecessary ground points by modeling the ground as a Markov Random Fields and applying belief propagation as it follows. 
First, we divide the point cloud according to a 2D grid of $0.4mx0.4m$ tiles on the $XY$ plane, where $X$ represents the forward direction, and $Y$ points to the left side of the moving vehicle. 
Starting from the cell at the origin of this grid, supposedly being ground, we move iteratively to the surrounding cells in order to propagate the ground height and to classify the tiles between ground and non-ground (see Fig. \ref{fig:propag}).
Let consider the cell $C_{ij}$ and the set $P_{ij}$ of the points projecting on this cell. We define $\hat{h}_G^{ij} = max\left\{h_G^N\right\}$ where $N$ is the set of neighboring cells, belonging to the inner ring, that propagate to  $C_{ij}$; then $H_{ij} = max\left\{P_{ij}^z\right\}$ and $h_{ij} = min\left\{P_{ij}^z\right\}$ are the maximum and minimum heights of the points in the cell (recall that coordinate $z$ represents the height of a point). 
Given a maximum expected slope of $22\%$ of the cell dimension, which is about $s=0.09m$; a cell is classified as ground plane if and only if:
\begin{equation}
H_{ij} - h_{ij} < s \qquad \text{and} \qquad H_{ij} < \hat{h}_G^{ij} + s.
\end{equation}
Then the current propagated ground height $h_G^{ij}$ is $H_{ij}$ if ${C}_{ij}$ is ground, otherwise it is $\hat{h}_G^{ij}$.
In Fig. \ref{fig:groundremoval} we illustrate an example of the ground points detected in a single scan.


\subsection{Moving Points detection}
After registration, point filtering, and ground removal we apply the laser-based moving object detection algorithm, which borrows some ideas from ~\cite{xiao2013change} and \cite{vallet2015extracting}. From the former we borrow the use of Dempster-Shafer Theory (DST) for occupancy space representation and the Dempster combination rule for intra-scan evidence fusion; from the latter we borrow the idea of using previous and future scans.

At first, we evaluate the occupancy of a point $P$ belonging to scan $S_{\text{k}}$ induced by another scan $S_{\text{i}}$ by representing the occupancy space using DST. 
The space occupancy is represented using a set $X = \{empty, occupied\}$; the DST operates on the power set of $X$, i.e., $2^X = \{\{\emptyset\}, \{empty\}, \{occupied\}, \{empty,occupied\}\}$, where the subset $\{empty,occupied\}$ represents the $unknown$ state, i.e., the space not reached by the beams. DST defines a degree of belief $m(\cdot)$ for each subset: for the empty set it is 0 and for the other subsets they are within the range of $[0, 1]$ and they add up to a total of 1.

\begin{figure}[t]
\centering
\includegraphics[width=0.7\columnwidth]{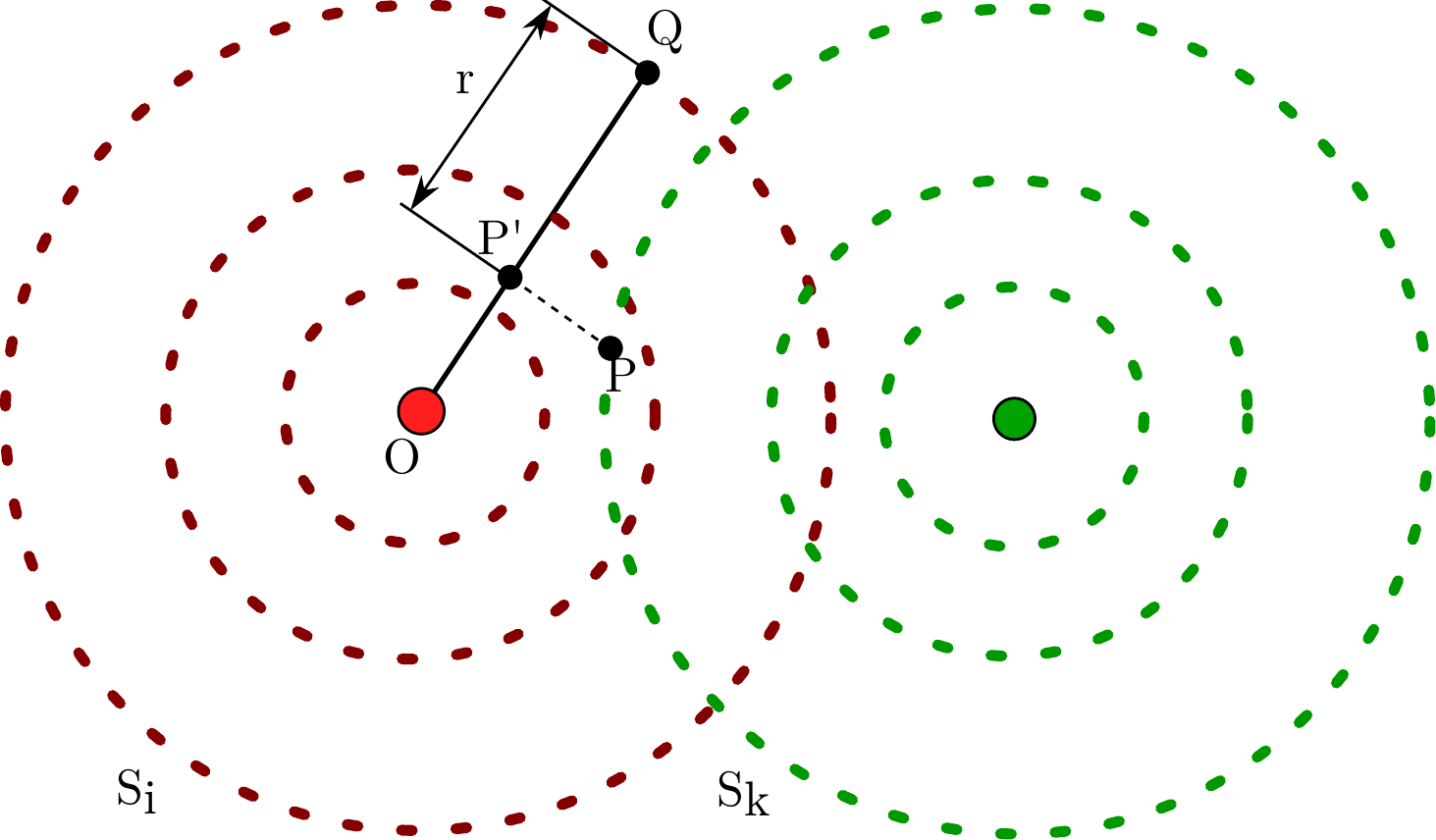}
\caption{Occupancy at point $P$ computed with respect to the beam $OQ$.}
\label{fig:scanocc}
\end{figure}

Let  $e = m(\{empty\})$, $o = m(\{occupied\})$ and $u = m(\{unknown\})$ be the degrees of belief for the three possible labels such that $e + o + u = 1$. 
Let $OQ$ be a laser beam of $S_i$, and $r = length(P'Q)$, where $P'$ is the projection of $P$ on $OQ$ (see Fig.~\ref{fig:scanocc}); then we define the degree of belief $e_r$ and $o_r$ parametrized over $r$ as it follows:
\begin{align}
\label{eq:occ}
 e_r &=  \begin{cases}
               1 \ \ & \text{if $Q$ is behind $P'$}\\
               0 \ \ & \text{otherwise}
         \end{cases},\\
\label{eq:occ_2}
o_r &=  \begin{cases}
			e^{-\frac{r^2}{2}} \ \ & \text{if $P'$ is behind $Q$}\\
			0 \ \ & \text{otherwise}
        \end{cases}.
\end{align}
The occupancy values at point $P$ due to the beam $OQ$ becomes then:
\begin{equation}
\label{eq:comb}
 m(P,Q)=\left\{\  \begin{aligned}
                 e\\o\\u
                 \end{aligned}
\  \right\} = \left\{\  \begin{aligned}
                     f_\theta&\cdot e_r \ \\
                     &o_r \ \\
                     1 - &e - o \ 
                    \end{aligned}
\ \right\}\\
\end{equation}
where $f_\theta = e^{-\frac{\theta^2}{2\lambda_\theta^2}}$ is the rotation occupancy function, $\lambda_\theta$ is the angular resolution of the sensor, and $\theta$ is the angle between rays $OP$ and $OQ$.

To embed uncertainty in this framework, we propose to model noise as a Gaussian variable, then we define $\sigma_m$, $\sigma_r$ and $\sigma_\theta$ as, respectively, measurement, registration, and angle standard deviations (with $\sigma_m=0.05 m$, $\sigma_r=0.15 m$, $\sigma_{\theta}=0.1\pi rad$). 
By defining  $g(m) = \mathcal{N}(0, \sigma_m^2)$, $g(r) = \mathcal{N}(0, \sigma_r^2)$, $F = g(m)\otimes g(r)$ , we modify \eqref{eq:comb} as it follows:

\begin{equation}
 m'(P,Q)=\left\{\  \begin{aligned}
                 e'\\o'\\u'
                 \end{aligned}
\  \right\} = \left\{\  \begin{aligned}
                     f_\theta  &\cdot (e_r \otimes F)\\
                     &o_r \otimes F\\
                     1 - &e' - o'\\ 
                    \end{aligned}
\ \right\}
\end{equation}
where $\otimes$ represents the convolution operator.
We  aggregate the occupancy induced by two beams through the Dempster combination rule applied to the occupancy induced by two beams with $m(P,Q_1)=(e_1, o_1, u_1)$ and $m(P,Q_2)=(e_2, o_2, u_2)$:

\begin{equation}
  \left\{ \begin{aligned}
                 e_1\  \\o_1\  \\u_1\ 
                \end{aligned}
 \right\} \oplus \left\{ \begin{aligned}
                     e_2\ \\o_2\ \\u_2\  
                    \end{aligned}
\right\} = \frac{1}{1-K} 
\left\{ 
  \begin{aligned}
     &e_1\cdot e_2 + e_1\cdot u_2 + u_1\cdot e_2 \ \\
     &o_1\cdot o_2 + o_1\cdot u_2 + u_1\cdot o_2 \ \\
     &u_1\cdot u_2 \ 
  \end{aligned}
 \right\}
\end{equation}
where $\oplus$ is the fusion operator defined by DST which is commutative and associative, and $K = o_1\cdot e_2 + e_1\cdot o_2$.
From this, the overall occupancy at location $P$ due to the $I$ neighboring rays $Q_i$ is then given by 
$m(P) = \bigoplus_{i\in I} m(P,Q_i)$.

To classify a point $P$ belonging to a scan $S_{\text{k}}$ as static or moving, we compute and combine its occupancy values due to previous and future\footnote{We use a time window of $2K$ scans around the current one, with $K=10$ in our experiments, introducing a K scans delay in the whole pipeline.} scans $\mathbb{S} = \{S_{\text{k-K}}, \dots, S_{\text{k-1}}, S_{\text{k+1}}, \dots, S_{\text{k+K}} \}$.
By comparing two scans having the degree of belief $m(P,Q_1)$ and $m(P,Q_2)$, a moving object corresponds to the not consistent degree of belief. To this extent, we compute:
\begin{equation}
 \begin{split}
  Conf &= e_1\cdot o_2 + o_1\cdot e_2\ \\
  Cons &= e_1\cdot e_2 + o_1\cdot o_2 + u_1\cdot u_2\ \\
  Unc\ &= u_1\cdot (e_2 + o_2) + u_2\cdot (e_1 + o_1)\\
 \end{split}
\end{equation}
where $Conf$ means conflicting, $Cons$ is consistent and $Unc$ uncertain. 
Moving points regions are those where $Conf > Cons$ and $Conf > Unc$. 
We have extended this procedure, originally proposed in~\cite{xiao2013change} for 2 scans, to compare $2K$ subsequent scans.
To do so, we propose to change the occupancy computation procedure in order to make the classification more robust by a novel discretized version of the original DST approach we just explained. 

Let consider the most distant point $B$ in each scan $S_{\text{i}} \in \mathbb{S}$, we approximate the occupancy values of $P$ with respect to $S_{\text{i}}$ in the following way. Let's define 
\[
 l = r_{sup} - \delta r\frac{||\overrightarrow{OP}||}{||\overrightarrow{OB}||}
\] 
where $r_{sup}$ and $r_{inf}$ are user defined upper and lower bounds and $\delta r = r_{sup} - r_{inf}$ (in our case  $r_{sup} = 0.8$ and $r_{inf} = 0.6$); $l$ is used to define a belief stronger in the neighborhood of the sensor. Then, from the original occupancy $m(P,Q)=(e, o, u)$ we derive the new occupancy of $P$ for any $Q\in S_{\text{i}}$:
\begin{equation}
 e_{\text{new}} = \begin{cases}
      l\ \ \ &\mbox{if }e > o \wedge e > u \\
      0 &\mbox{otherwise}
     \end{cases}
\end{equation}

\begin{equation}
 o_{\text{new}} = \begin{cases}
      l\ \ \ &\mbox{if }o > e \wedge o > u \\
      0 &\mbox{otherwise}
     \end{cases}
\end{equation}

\begin{equation}
 u_{\text{new}} = 1 - e_{\text{new}} - o_{\text{new}}.
\end{equation}

 This way the occupancy value of each point is discretized based on its distance from the sample scan origin. With these discretized values we apply again the Dempster combination rule among the set $\mathbb{S}$ of scans, and the outcome of this combination defines the classification of the point: if its prevalent occupancy state is $empty$ then the point is considered to be dynamic, otherwise it is a static point.

Testing every point $P$ from a scan $S_{\text{k}}$ against every neighboring ray in the $\mathbb{S}$ scans is a very expensive procedure. 
We avoid such expensive computations by indexing the $S_{\text{k}}$ point cloud with an octree data structure, with a resolution of 0.3m, and we perform the tests only for a small set of points in its nodes, i.e., in the neighborhood of the point $P$.
Since dynamic points in real world are not sparse, as they are part of a moving object, we assume they have neighboring dynamic points, and, if a small set of neighboring points are classified as dynamic, their neighbors should also be considered dynamic as well. 
Thus, to improve the performance of our algorithm, for each leaf of the octree, we perform the moving object detection test on a random subset of points ($\frac{1}{6}$ of the total amount) and if there are at least half of the tested points classified as dynamic, then we classify all the points in the leaf as such. Otherwise, the leaf is assumed to contain only static points.
If the number of points in a leaf is small, i.e., less than $\tau_{np}=6$, they are sparse and they all get tested. By doing this way, we do not only improve the computational efficiency of the algorithm, but we also reduces the amount of misclassified dynamic points in static objects.

\section{Image-based Moving Object Validation}
\label{sec:images}
Even if the outcome of our laser-based moving object detection algorithm is often satisfying, some false positives may arise due to the noise in the laser measurements and the inaccuracy of the point cloud registration step. 
We propose thus two additional image-based validation tests to filter out false positives: both tests compare image patches around the projection of 3D points classified as moving objects respectively in the (color) images corresponding to the scans in $\mathbb{S}$ and in the depth-maps estimated from the point clouds themselves. If a candidate point passes these tests, then it is confirmed to be a moving point.

\begin{figure}[t]
\centering
\includegraphics[width=0.8\columnwidth]{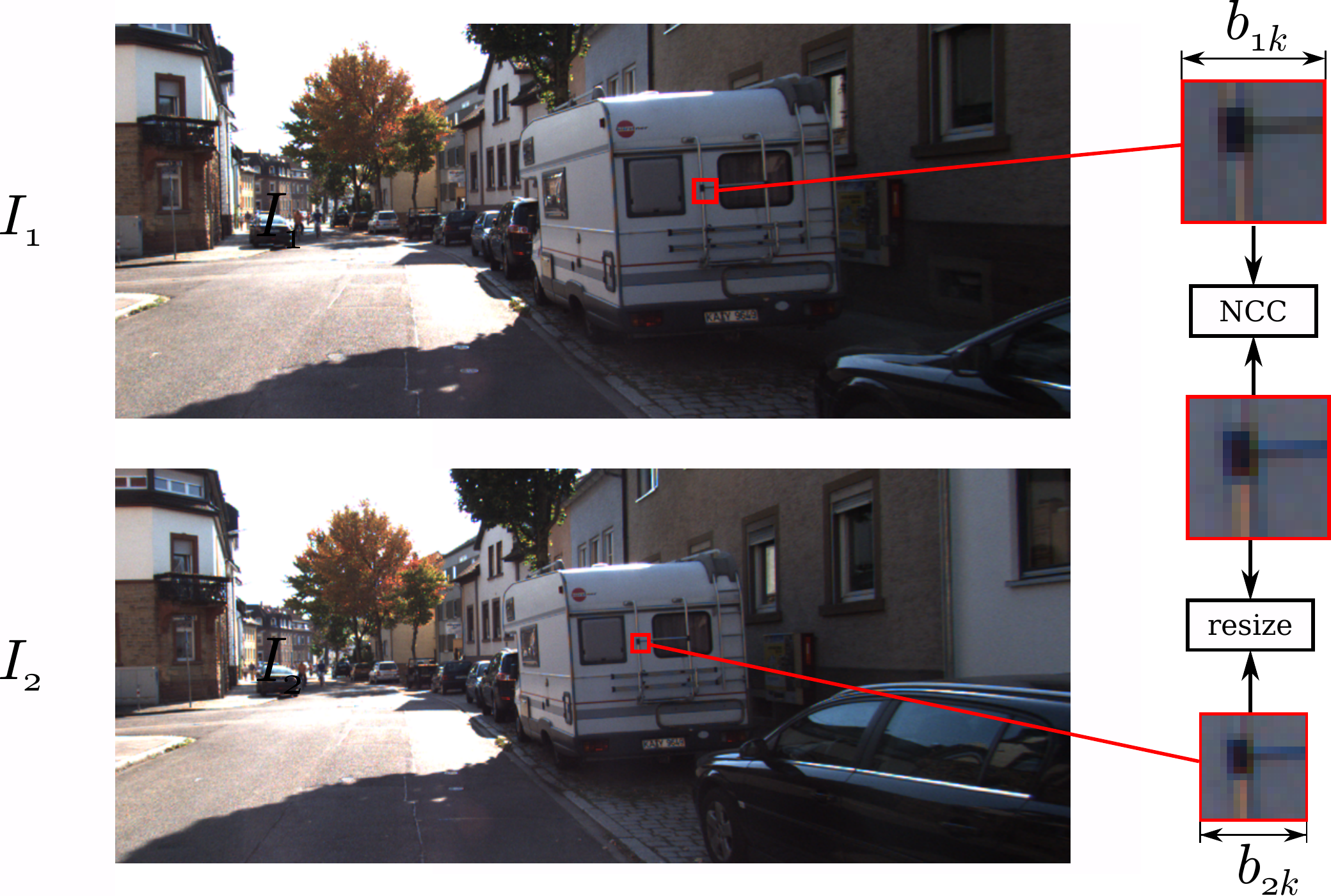}
\caption{Example of two color patches compared via NCC.}
\label{fig:ncc}
\end{figure}

In the proposed tests a 3D point $P_i$ is projected into a pixel $p_{ik}$ of the (color) image $I_k$ and depth-map $D_k$ by using the camera calibration matrix. A squared image patch $patch_{ik}$ around this pixel is selected having a side length $b_{ik}$, measured in number of pixels, according to the following formula:
\begin{equation}
 b_{ik} = \frac{h}{d_{ik}}f_{xy},
\end{equation}
where the $h=0.15m$ parameter refers to the patch height in the real world, $d_{ik}$ is the distance of the point $P_i$ from the camera $k$ corresponding to pixel $p_{ik},$ and $f_{xy}$ is the focal length of the camera (see Fig. \ref{fig:ncc}).
Since each comparison needs two patches of the same size, one of the two patches, in turn, is resized.

Before computing any similarity between images patches, these are checked for uniformity in their intensities. If their intensity standard deviation is above a certain threshold (0.02 in our case), then we compare the color patches, otherwise, the test would fail, so we compare the depth-maps.

\subsection{Image patch test}
Once patches are extracted and resized, we test their similarity through Normalized Cross Correlation (NCC) on each color channel independently. 
Let $\pi_i$ and $\pi_j$ be the two patches and $NCC_c(u,v)$ the NCC between two images at location $(u,v)$ computed for the $c$-th channel, then we define:
\begin{equation}
 E_c(\pi_i, \pi_j) = 1 - \max_{u,v}NCC_c(u,v).
\end{equation}
Two patches are considered to be similar if, for all cameras in $\mathbb{S}$, $E_c < \tau$ (in our case $\tau=0.1$). 

We opted for NCC measure with respect to the classical Sum of Squared Differences (SSD), since it is not affected by illumination changes issue, even thought it is computationally more expensive than SSD. Note that, if one of the patches has one of its channels flat, the formula above fails, as we get a division by zero problem for the NCC computation. Our system handles this case with the test on the depth-maps. 

\subsection{Depth-map patch test}
\begin{figure}[t]
\centering
\includegraphics[width=0.8\columnwidth]{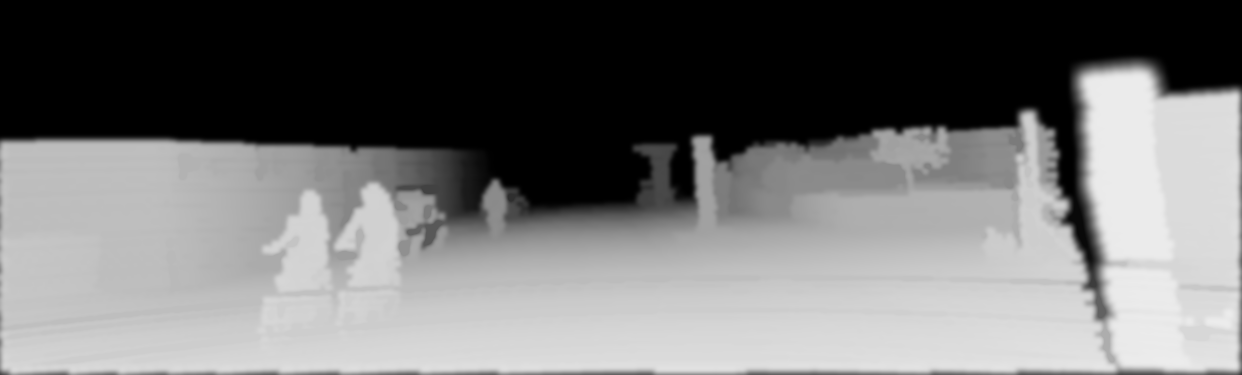}
\caption{Example of a depth-map computed from a lidar point cloud.}
\label{fig:depth}
\end{figure}

When the color image patch test fails, we apply a comparison between the depth-maps extracted from the lidar data as it follows.
We project lidar points into the $k$-th camera, by using the camera matrix, and we create a sparse depth-map having for each pixel the camera-to-point distance. Then we apply a disk shaped dilation such that we close all the gaps between close points.
The resulting image is a rough estimation of the depth-map of the lidar points, nevertheless its computation is fast and the result has been sufficiently discriminative in our tests (Fig. \ref{fig:depth} shows an example of a depth-map extracted from the laser data).

Since the laser sensor is moving between the two scans, we need to correct the depth-maps for this movement before we can actually compare the selected patches.
To perform this correction, we assume a small motion between two cameras and we use the following formula:
\begin{equation}
D_l' = D_l + \frac{||\mathbf{t}_k - \mathbf{t}_l||}{d_{l,max}}
\end{equation}
where $D_l'$ and $D_l$ are respectively the new and old intensity values of depth-map $l$, $\mathbf{t}_k$ and $\mathbf{t}_l$ are translation vectors, with respect to a global reference frame, for cameras $k$ and $l$ respectively, and $d_{l,max}$ is the maximum depth distance from the camera $l$.

Then patches are extracted and resized the same way we do for (color) images, but in the depth-map comparison we use the SSD metric. This metric is suitable in this case because depth-maps are not affected by illumination changes.

\section{Experimental results}
\label{sec:experiments}
\begin{figure}[t]
\centering
\includegraphics[width=0.8\columnwidth]{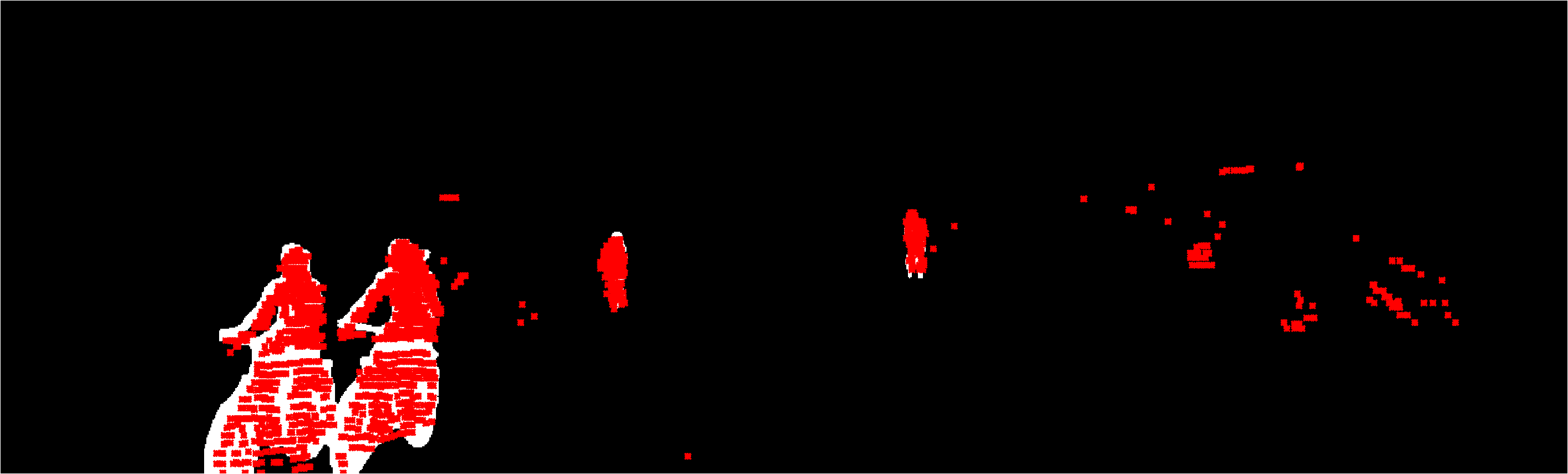}
\caption{Points classified as moving projected against the ground-truth mask.} 
\label{fig:points}
\end{figure}

To the best our knowledge a dataset with surveying camera having annotated moving objects is not available, so we tested the proposed algorithm with three sequences of the KITTI~\cite{Kitti} dataset, which provides 1392x512px images, camera calibration information, and Velodyne HDL-64E point clouds, where we manually annotated the moving object regions on the images. To evaluate the accuracy of the classification, we project the 3D points of each point cloud on the corresponding image plane and we check if points classified as dynamic objects project into the manually annotated masks. An example of the comparison between the resulting dynamic points and ground truth mask is shown in Fig.~\ref{fig:points}.
We run the tests on a  Intel Core i7-3537u (2 Cores), 2GHz with 8GB of DDR3 RAM. 

We compare our approach with the state-of-the-art Vallet et al. \cite{vallet2015extracting} which is the approach closer to the proposed.
In Table \ref{tab:resPR} we list the precision/recall results; our laser-based algorithm, by paying a very small decrease in recall, it increases significantly the precision of \cite{vallet2015extracting}, and image validation refines the results further.
In Table \ref{tab:numPRobj} we show that the proposed algorithm detects or partially detects a higher number of moving object. Here  partially detected means that a subset of the moving points remains in the final global cloud.

\begin{figure}[t]
\centering
\includegraphics[width=0.8\columnwidth]{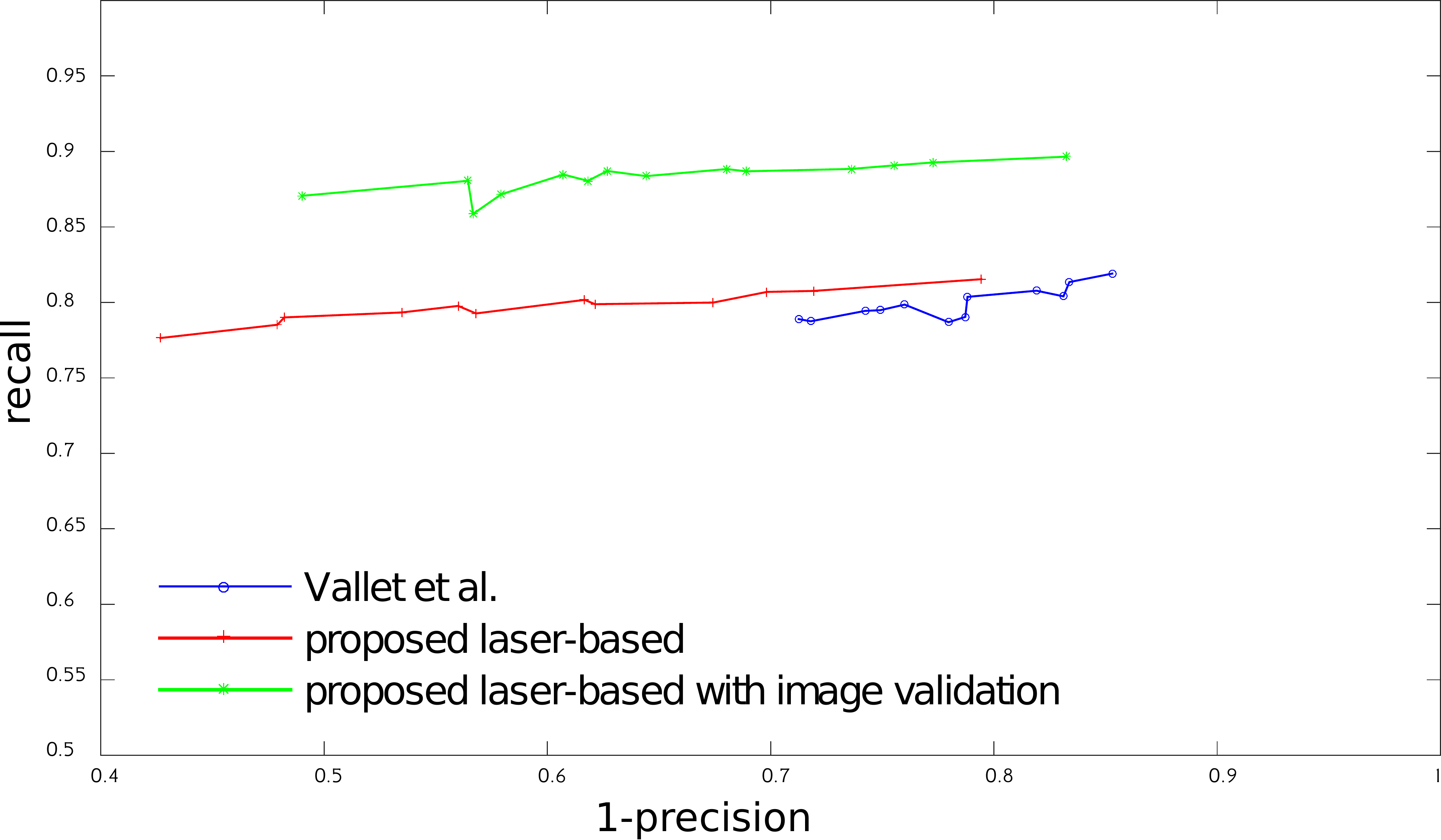}
\caption{ROC of ~\cite{vallet2015extracting} and our algorithms.}
\label{fig:roc}
\end{figure}

In Fig.~\ref{fig:roc} we report the Receiver Operating Characteristics (ROC) curve obtained with the 0095 sequence of the KITTI dataset: here we compare the ground truth mask against an image-based mask of moving objects obtained with a simple dilation of the points classified as moving and projected in the image plane, to have a result similar to classic background subtraction algorithms. 
In the ROC curve, the highest the area subtended by the curve, the better the classifier performance; precision and recall reported in the plot are obtained by varying the $\sigma_r$ and $\sigma_{\theta}$ parameters such that $0.1m<\sigma_r<0.45m$ and $0.0035rad<\sigma_{\theta}<0.0088rad$.
The results enforce the conclusion that the proposed approach performs better that the algorithm by Vallet et al. already in the lidar-based only version and shows the overall improvement with the image-based validation.
The novel laser-based pipeline and the image validation procedure improve significantly the precision of the proposed algorithm, in particular the validation has been able to discard a huge number of false positive in moving objects detection.
Discretization and diffusion of occupancy information lead to smoother and more precise results.

Our algorithm outperforms the work by Vallet et al. also in terms of computing speed, thanks to the use of octree indexing and subsampling. 
Indeed, our algorithm takes, on average, 0.6 seconds per point cloud, while Vallet et al. approach takes 4.9 seconds.
Timing does not include the image-based validation step, this has been implemented as a prototype in MATLAB and, at the current stage, it works off-line at 25 seconds per frame to estimate the depth-map and it requires 1-2 seconds to validate the moving points. Nevertheless, this step can be easily parallelized on GPU leading to real time computations.

\begin{table}[t]
\caption{Precision/recall results on KITTI sequences.}
\label{tab:resPR}
\centering
\setlength{\tabcolsep}{3px}
\begin{tabular}{lcccccc}
\toprule 										
&\multicolumn{2}{c}{seq 0091}&\multicolumn{2}{c}{seq 0095}&\multicolumn{2}{c}{seq 0104}\\
&P & R &P & R &P & R \\
\midrule
Vallet el al.~\cite{vallet2015extracting}	&	0.11&\textbf{0.80}	&	0.10&\textbf{0.81}	& 0.27&\textbf{0.90}\\
Proposed laser-based                		&   0.25&0.75 	&  0.19&0.79  	& 0.44&0.87\\
Proposed w/ image validation 				&  	\textbf{0.26}&0.73	&  \textbf{0.24}&0.76  	& \textbf{0.49}&0.89\\
\end{tabular}
\end{table}

\begin{table}[t]
\caption{Number of  moving objects Detected (D) and partially detected (PD).}
\label{tab:numPRobj}
\centering
\begin{tabular}{lcccccc}
\toprule 
&\multicolumn{2}{c}{seq 0091}&\multicolumn{2}{c}{seq 0095}&\multicolumn{2}{c}{seq 0104}\\
&D & PD &D & PD &D & PD \\
\midrule
Vallet el al.~\cite{vallet2015extracting}	& 6/20 	& 7/20	& 8/8 &   0/8	&    1/6	& 3/6\\
Proposed & \textbf{8/20}	& \textbf{9/20}	& 8/8 &   0/8	&    \textbf{3/6}	& 3/6 \\
\end{tabular}
\end{table}

\section{Conclusion and Future Work}
\label{sec:concl}
In this paper we propose a novel moving objects detection algorithm which improves the current state-of-the-art in laser-based approaches. The proposed approach relies on Demspter-Shafer Theory to model the occupancy induced by the data in a point cloud and to detect which point is dynamic or static. Moreover, we added a novel image validation step to remove false positive detections.
Experiments show that ground plane removal and scan comparison discretization improve on precision with respect to current state-of-the-art with a speed-up in the execution thanks to the use of an efficient indexing data structure.
As a future work we aim at applying the proposed approach to an existing urban reconstruction method~\cite{romanoni2015incremental} and refine it with ~\cite{romanoni16} in order to obtain a 3D urban map without moving objects, while still improving on the speed up of the visual pipeline of our proposal.

\section*{Acknowledgments}
{\small
This work has been supported by the POLISOCIAL Grant ``Maps for Easy Paths (MEP)'', the ``Interaction between Driver Road Infrastructure Vehicle and Environment (I.DRIVE)'' Inter-department Laboratory Grant from Politecnico di Milano and Nvidia who kindly support our research through the  Hardware Grant Program. }

\bibliographystyle{IEEEtran}
\bibliography{IEEEabrv,biblio}

\begin{thebibliography}{10}
\providecommand{\url}[1]{#1}
\csname url@samestyle\endcsname
\providecommand{\newblock}{\relax}
\providecommand{\bibinfo}[2]{#2}
\providecommand{\BIBentrySTDinterwordspacing}{\spaceskip=0pt\relax}
\providecommand{\BIBentryALTinterwordstretchfactor}{4}
\providecommand{\BIBentryALTinterwordspacing}{\spaceskip=\fontdimen2\font plus
\BIBentryALTinterwordstretchfactor\fontdimen3\font minus
  \fontdimen4\font\relax}
\providecommand{\BIBforeignlanguage}[2]{{%
\expandafter\ifx\csname l@#1\endcsname\relax
\typeout{** WARNING: IEEEtran.bst: No hyphenation pattern has been}%
\typeout{** loaded for the language `#1'. Using the pattern for}%
\typeout{** the default language instead.}%
\else
\language=\csname l@#1\endcsname
\fi
#2}}
\providecommand{\BIBdecl}{\relax}
\BIBdecl

\bibitem{azim2012detection}
A.~Azim and O.~Aycard, ``Detection, classification and tracking of moving
  objects in a 3d environment,'' in \emph{Intelligent Vehicles Symposium (IV),
  2012 IEEE}.\hskip 1em plus 0.5em minus 0.4em\relax IEEE, 2012, pp. 802--807.

\bibitem{vieira2014spatial}
A.~W. Vieira, P.~L. Drews, and M.~F. Campos, ``Spatial density patterns for
  efficient change detection in 3d environment for autonomous surveillance
  robots,'' \emph{Automation Science and Engineering, IEEE Transactions on},
  vol.~11, no.~3, pp. 766--774, 2014.

\bibitem{andreasson2007has}
H.~Andreasson, M.~Magnusson, and A.~Lilienthal, ``Has somethong changed here?
  autonomous difference detection for security patrol robots,'' in
  \emph{Intelligent Robots and Systems, 2007. IROS 2007. IEEE/RSJ International
  Conf. on}.\hskip 1em plus 0.5em minus 0.4em\relax IEEE, 2007, pp. 3429--3435.

\bibitem{drews2013fast}
P.~Drews, S.~da~Silva~Filho, L.~Marcolino, and P.~N{\'u}nez, ``Fast and
  adaptive 3d change detection algorithm for autonomous robots based on
  gaussian mixture models,'' in \emph{Robotics and Automation (ICRA), 2013 IEEE
  International Conf. on}.\hskip 1em plus 0.5em minus 0.4em\relax IEEE, 2013,
  pp. 4685--4690.

\bibitem{xiao2013change}
W.~Xiao, B.~Vallet, and N.~Paparoditis, ``Change detection in 3d point clouds
  acquired by a mobile mapping system,'' \emph{ISPRS Annals of Photogrammetry,
  Remote Sensing and Spatial Information Sciences}, vol.~1, no.~2, pp.
  331--336, 2013.

\bibitem{nunez2010change}
P.~N{\'u}{\~n}ez, P.~Drews~Jr, A.~Bandera, R.~Rocha, M.~Campos, and J.~Dias,
  ``Change detection in 3d environments based on gaussian mixture model and
  robust structural matching for autonomous robotic applications,'' in
  \emph{Intelligent Robots and Systems (IROS), 2010 IEEE/RSJ International
  Conf. on}.\hskip 1em plus 0.5em minus 0.4em\relax IEEE, 2010, pp. 2633--2638.

\bibitem{Piccardi2004background}
M.~Piccardi, ``Background subtraction techniques: a review,'' in \emph{Systems,
  man and cybernetics, 2004 IEEE international conference on}, vol.~4.\hskip
  1em plus 0.5em minus 0.4em\relax IEEE, 2004, pp. 3099--3104.

\bibitem{Sobral2014}
A.~Sobral and A.~Vacavant, ``{A comprehensive review of background subtraction
  algorithms evaluated with synthetic and real videos.}'' \emph{Computer Vision
  and Image Understanding}, vol. 122, p. 4–21, 2014.

\bibitem{azzari2005effective}
P.~Azzari, L.~D. Stefano, and A.~Bevilacqua, ``An effective real-time mosaicing
  algorithm apt to detect motion through background subtraction using a ptz
  camera,'' in \emph{Advanced Video and Signal Based Surveillance, 2005. AVSS
  2005. IEEE Conf. on}.\hskip 1em plus 0.5em minus 0.4em\relax IEEE, 2005, pp.
  511--516.

\bibitem{Andrea}
A.~Romanoni, M.~Matteucci, and D.~G. Sorrenti, ``Background subtraction by
  combining temporal and spatio-temporal histograms in the presence of camera
  movement,'' \emph{Machine vision and applications}, vol.~25, no.~6, pp.
  1573--1584, 2014.

\bibitem{kim2013detection}
S.~W. Kim, K.~Yun, K.~M. Yi, S.~J. Kim, and J.~Y. Choi, ``Detection of moving
  objects with a moving camera using non-panoramic background model,''
  \emph{Machine vision and applications}, vol.~24, no.~5, pp. 1015--1028, 2013.

\bibitem{shakeri2014detection}
M.~Shakeri and H.~Zhang, ``Detection of small moving objects using a moving
  camera,'' in \emph{Intelligent Robots and Systems (IROS 2014), 2014 IEEE/RSJ
  International Conf. on}.\hskip 1em plus 0.5em minus 0.4em\relax IEEE, 2014,
  pp. 2777--2782.

\bibitem{markovic2014moving}
I.~Markovic, F.~Chaumette, and I.~Petrovic, ``Moving object detection, tracking
  and following using an omnidirectional camera on a mobile robot,'' in
  \emph{Robotics and Automation (ICRA), 2014 IEEE International Conf.
  on}.\hskip 1em plus 0.5em minus 0.4em\relax IEEE, 2014, pp. 5630--5635.

\bibitem{lin2014deep}
T.-H. Lin and C.-C. Wang, ``Deep learning of spatio-temporal features with
  geometric-based moving point detection for motion segmentation,'' in
  \emph{Robotics and Automation (ICRA), 2014 IEEE International Conf.
  on}.\hskip 1em plus 0.5em minus 0.4em\relax IEEE, 2014, pp. 3058--3065.

\bibitem{premebida2009lidar}
C.~Premebida, O.~Ludwig, and U.~Nunes, ``Lidar and vision-based pedestrian
  detection system,'' \emph{Journal of Field Robotics}, vol.~26, no.~9, pp.
  696--711, 2009.

\bibitem{vallet2015extracting}
B.~Vallet, W.~Xiao, and M.~Br{\'e}dif, ``Extracting mobile objects in images
  using a velodyne lidar point cloud,'' \emph{ISPRS Annals of Photogrammetry,
  Remote Sensing and Spatial Information Sciences}, vol.~1, pp. 247--253, 2015.

\bibitem{Kitti}
A.~Geiger, P.~Lenz, C.~Stiller, and R.~Urtasun, ``Vision meets robotics: The
  kitti dataset,'' \emph{International Journal of Robotics Research (IJRR)},
  2013.

\bibitem{segal2009generalized}
A.~Segal, D.~Haehnel, and S.~Thrun, ``Generalized-icp.'' in \emph{Robotics:
  Science and Systems}, vol.~2, no.~4, 2009.

\bibitem{romanoni2015incremental}
A.~Romanoni and M.~Matteucci, ``Incremental reconstruction of urban
  environments by edge-points delaunay triangulation,'' in \emph{Intelligent
  Robots and Systems (IROS), 2015 IEEE/RSJ International Conference on}, Sept
  2015, pp. 4473--4479.

\bibitem{romanoni16}
A.~Romanoni, A.~Delaunoy, M.~Pollefeys, and M.~Matteucci, ``Automatic 3d
  reconstruction of manifold meshes via delaunay triangulation and mesh
  sweeping,'' in \emph{2016 IEEE Winter Conference on Applications of Computer
  Vision (WACV)}, March 2016, pp. 1--8.

\end{thebibliography}
\end{document}